\documentclass[11pt,a4paper]{article}
\usepackage{times,latexsym}
\usepackage{url}
\usepackage[T1]{fontenc}

\usepackage[acceptedWithA]{tacl2021v1}
\usepackage{xspace,mfirstuc,tabulary}

\usepackage{helvet}
\usepackage{courier}
\usepackage{graphicx}

\usepackage{booktabs} % For tables
\usepackage{blindtext}
\usepackage{algorithm} % For pseudocode
\usepackage[noend]{algpseudocode} % For pseudocode
\usepackage{tikz} % For plate notation graph
\usepackage{setspace} % For adjusting line spacing
\usepackage{color}
\usepackage{amsmath}
\usepackage{amssymb}
\usepackage{amsthm}

\usepackage{enumerate}
\usepackage{enumitem}
\usepackage{multirow}

\usepackage{esvect}

\usepackage{color,soul}

\usepackage{caption}
\usepackage{subcaption}
\usepackage{tabularx}
\usepackage{makecell}

\title{Active entailment encoding for explanation tree construction using parsimonious generation of hard negatives}

\author{
  Alex Bogatu\Thanks{The first two authors contributed equally.}~ $^\diamond$ $^\dagger$ , 
  Zili Zhou\footnotemark[1]~ $^\diamond$ $^\dagger$ ,
  D\'{o}nal Landers $^\dagger$ ,
  Andr\'{e} Freitas $^\diamond$ $^\dagger$ $^\ddagger$
  \\
  \ \\
  $^\diamond$ Department of Computer Science, The University of Manchester
  \ \\
  $^\dagger$digital Experimental Cancer Medicine Team
  \\
  Cancer Biomarker Centre, CRUK Manchester Institute
  \ \\
  $^\ddagger$ Idiap Research Institute, Switzerland
  \ \\
}

\date{}

\begin{document}
\maketitle
\begin{abstract}

Entailment trees have been proposed to simulate the human reasoning process of explanation generation in the context of open--domain textual question answering. However, in practice, manually constructing these explanation trees proves a laborious process that requires active human involvement. Given the complexity of capturing the line of reasoning from question to the answer or from claim to premises, the issue arises of how to assist the user in efficiently constructing multi--level entailment trees given a large set of available facts. In this paper, we frame the construction of entailment trees as a sequence of active premise selection steps, i.e., for each intermediate node in an explanation tree, the expert needs to annotate positive and negative examples of premise facts from a large candidate list. We then iteratively fine--tune pre--trained Transformer models with the resulting positive and tightly controlled negative samples and aim to balance the encoding of semantic relationships and explanatory entailment relationships. Experimental evaluation confirms the measurable efficiency gains of the proposed active fine--tuning method in facilitating entailment trees construction: up to 20\% improvement in explanatory premise selection when compared against several alternatives.
\end{abstract}

\section{Introduction} \label{sec:intro}

\begin{figure}[t]
\centering
\includegraphics[width=1\linewidth]{"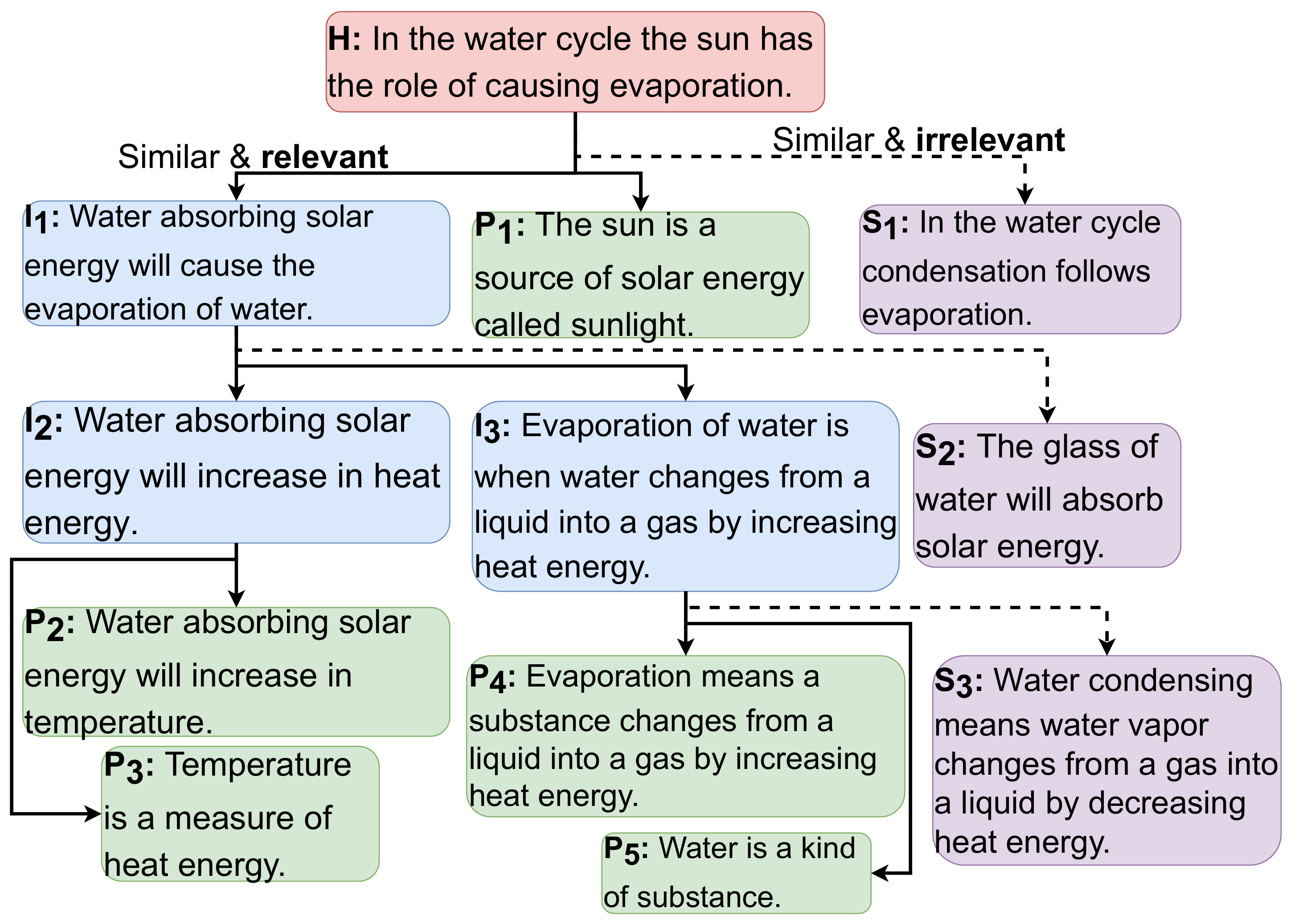"}
\caption{Entailment Bank sub--tree showing intermediate $I_j$ and leaf $P_i$ nodes relevant for explaining $H$, together with irrelevant but similar premises $S_k$.}
\label{fig:tree}
\end{figure}

Multi--level entailment trees have been recently proposed to instill reasoning capabilities based on complex explanation chains into Transformer--based models for explanation generation in the context of question answering (QA) and textual entailment (e.g., Entailment Bank \cite{Dalvi2021ExplainingAW}). Models fine--tuned on such structures could lead to more expressive explanations than retrieval--based (e.g., \cite{deyoung-etal-2020-eraser} or multi--hop (e.g., \cite{JhamtaniC20, Valentino22} alternatives. This expressiveness comes from a core property of entailment trees: to describe at each level a diverse set of multi--premise textual entailment relationships that link a hypothesis (i.e., an answer to a question) can be explained from available textual evidence. However, in practice there is a scarcity of such tree benchmarks and constructing entailment trees from a given textual corpus is a labor intensive task that requires active user involvement \cite{Dalvi2021ExplainingAW}. A practical way to reduce the expert's workload is to employ retrieval--based models (e.g., \cite{Karpukhin2020DensePR}) iteratively. When used, these models take root and intermediate tree nodes as input and aim to return explanatory premises from the corpus. In practice, such an approach often exhibits \textit{similarity bias}, i.e., the retrieval of premise candidates that, although semantically or lexically similar to the query, do not contribute to the explanatory inference chain. Furthermore, more abstract explanatory statements with lower lexical and semantic overlap often missing from the result set of such retrieval models. 

As an example, consider the Entailment Bank \cite{Dalvi2021ExplainingAW} sub--tree illustrated in Figure \ref{fig:tree}. Constructing the non--dotted branches iteratively using pre--trained sentence/passage retrieval models involves treating each intermediate node $I_j$ as a query and selecting from the returned candidates the premises $P_i$ that are relevant for the intended explanation at each step. In practice, premises that share few or no words with their parent node $I_j/H$, e.g., $P_1$, $P_3$, may be missing from the top--K retrieved results. They are often replaced by lexically overlapping sentences that are irrelevant for the explanation. For example, premises $S_1, S_2, S_3$ on the dotted branches are among the retrieved results for $H$, $I_1$ and $I_3$, respectively. This type of behavior, called \textit{sentence similarity bias} in this paper, has also been observed in other works that study the construction of entailment trees, such as \cite{Ribeiro22}.

Our goal in this paper is, therefore, to reduce the bias of premise selection models toward semantic relevance and to balance it with entailment relevance for the purpose of iterative entailment tree construction. To this end, we fine--tune BERT--based \cite{Devlin2019BERTPO} bi--encoder models \cite{reimers-2019-sentence-bert} using a novel active learning \cite{burr-2009} strategy that can assist the user in both carefully selecting positive and negative examples for fine--tuning the models and generating tree branches when the models are already trained. Negative sampling has been shown to have a significant impact in contrastive learning \cite{Hadsell06-contrastive} scenarios, with \textit{hard negatives} being able to increase the effectiveness of contrastive models \cite{Zhang2021UnderstandingHN}. Our simple yet efficient active learning proposal selects such hard negative examples for premise selection, alongside positive examples. Since our ultimate goal is to facilitate the capture of explanatory entailment relations in the sentence/passage embeddings, we call our proposed method \textit{AE--Enc} for Active Entailment Encoding. Specifically, the main contributions of this paper are:

\begin{enumerate}[leftmargin=*, nosep]
    \item An empirical analysis of the \textit{sentence similarity bias} concept: a characteristic of many Transformer--based retrieval models that tend to favor semantic and lexical similarity at the expense of entailment relationships.
    \item A novel data sampling strategy, called \textit{Active Contrastive Sampling} (ACS), that allows for parsimonious selection of fine--tuning negative samples (i.e., hard negatives \cite{gillick-etal-2019-learning}) that mitigate the above mentioned bias.
    \item An empirical evaluation of an active fine--tuning process, AE--Enc, that relies on the above sampling strategy and that shows up to 20\% improvement in retrieval models' capacity of selecting explanation relevant premises when compared against multiple alternatives.
\end{enumerate}

\section{Related work}

The concept of natural language explanation has been firstly introduced in \cite{camburu2018snli}, while \cite{Rajani2019ExplainYL} used the idea on commonsense reasoning tasks. However, these works limit themselves at using only single textual descriptions for each explanation. Abstractive explanatory tasks have been proposed as a more principled methodology to evaluate deeper explanation capabilities \cite{thayaparan2020survey,xie2020worldtree,jansen2018worldtree}. Most of these more recent works in the area of explanation generation rely on the use of Transformer models \cite{vaswani2017attention}, although simpler alternatives such as the relevance--unification model \cite{valentino2020unification} achieved comparable results. \cite{JhamtaniC20} proposed Transformer--based multi--hop QA to extend single textual description to a chain of reasoning facts as explanation. Similarly, \cite{Valentino22} presented a hybrid autoregressive model for multi--hop abductive natural language inference. Further work have shown how such reasoning chains can be semantically constrained using a Linear Programming (LP) paradigm \cite{thayaparan2021explainable}.

\cite{Dalvi2021ExplainingAW} introduced a more detailed and expressive form of explanation in the form of entailment trees. These structures, although consistent with the multi--hop reasoning approaches mentioned above, often include more complex explanatory relationships grounded on explanatory entailment between claim and premises. Such relationships are often not synonymous with lexical and semantic similarity relationships that are ubiquitous in datasets used to pre--train Transformer--based solutions. \cite{Tafjord2021ProofWriterGI} aimed to generate such entailment trees using encoder--decoder models based on a limited set of supporting facts to choose from in generating explanations. \cite{Ribeiro22} combine retrieval and generative methods for generating explanation trees iteratively. Their approach achieves comparable results to the $T5$--based \cite{Raffel-T519} alternative proposed alongside the Entailment Bank dataset by \cite{Dalvi2021ExplainingAW}. However, none of these approaches manage to mitigate the sentence similarity bias of the common pre--trained Transformer models, such as EMLo \cite{Peters2018DeepCW}, Sentence-BERT \cite{reimers-2019-sentence-bert}, DPR \cite{Karpukhin2020DensePR} or SimCSE \cite{Gao2021SimCSESC}. In this paper, we aim to reduce this behavior by splitting the explanation tree generation into iterative steps and involving the user in parsimoniously selecting positive and negative examples of explanatory and non--explanatory premises at each level of the tree. Our proposed approach is aligned to the hard negative mining strategy proposed in \cite{gillick-etal-2019-learning} for entity linking. However, our work is centered around user--controlled sampling with the aim to generate fine--tuning data for explanatory entailment.

\section{Active Entailment Encoding for explanation tree generation}

We construe the task of generating explanation trees as an iterative premise selection problem. More formally, given a hypothesis $H$ to explain and a corpus $\mathcal{C}$ containing premise sentences or passages, the objective is to generate an entailment tree $T$ (e.g., such as the one illustrated in Figure \ref{fig:tree}) that combines explanatory--relevant premises from a subset of $\mathcal{C}$ to explain $H$. Ultimately, the tree structure aims to describe the chain of reasoning that leads to explaining $H$.

In this paper, we distinguish between three types of entailment tree nodes for $T$: hypothesis node $H$ - the root of the tree (e.g., in red in Figure \ref{fig:tree}), intermediate nodes $I_j$ - nodes that explain other intermediate nodes and/or the hypothesis and are themselves explained by other nodes (e.g., in blue in Figure \ref{fig:tree}), and leaf nodes $P_i$ - premises that are not further explained (e.g., in green in Figure \ref{fig:tree}). In addition, we consider spurious nodes $S_k$ that denote premises lexically or semantically similar to their parent $H$ or $I_j$ but irrelevant for the purpose of explanation (e.g., in purple in Figure \ref{fig:tree}). 

\begin{figure}[t]
\centering
\includegraphics[width=1\linewidth]{"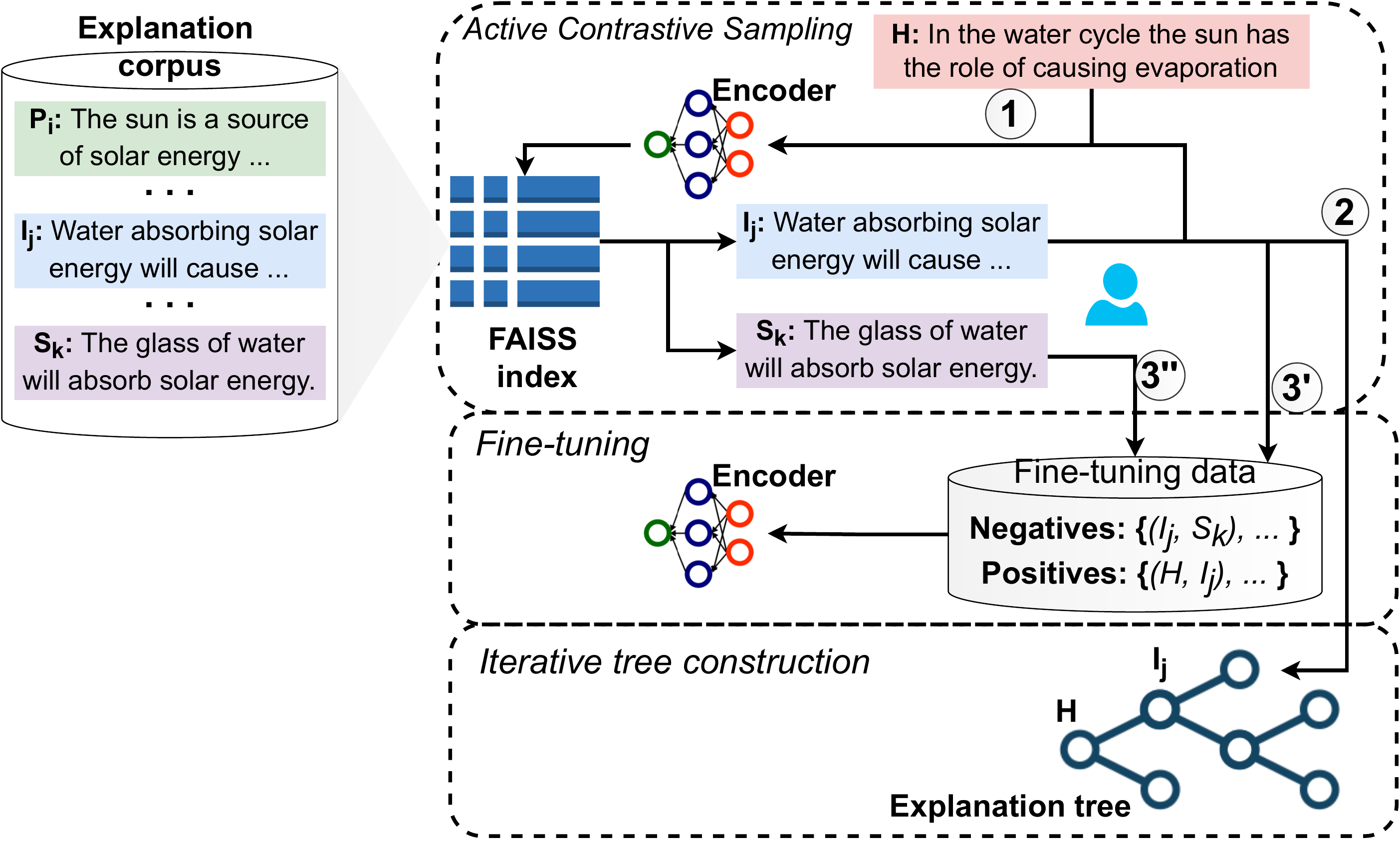"}
\caption{Active Entailment Encoding using parsimonious selection of positives/hard negatives for fine-tuning.}
\label{fig:approach}
\end{figure}

Figure \ref{fig:approach} broadly depicts a three--step approach for active generation of entailment trees that we call AE--Enc for Active Entailment Encoding. In designing it, we rely on the use of retrieval models, such as Sentence--BERT (SBERT) proposed in \cite{reimers-2019-sentence-bert} for sentence encodings or Deep Passage Retrieval (DPR) proposed in \cite{Karpukhin2020DensePR} for passage encodings. These encoder models are used to encode each premise in $\mathcal{C}$. The resulting embeddings are indexed into an efficient search data structure, such as FAISS \cite{Johnson-FAISS-21}. Then, starting from each hypothesis to explain $H$, we aim to build an explanation tree depth-first by iteratively retrieving the FAISS--neighbours of each $H$ and intermediate nodes $I_j$ (e.g., step $1$ in Figure \ref{fig:approach}). We assume this process is user--controlled, i.e., the user selects which of the retrieved candidates are relevant for explaining the query at each step (e.g., step $2$ in Figure \ref{fig:approach}). Therefore, the top--K retrieval process needs to return relevant candidates so that the user can identify explanatory premises. Since the pre--trained encoder can lead to sentrnce similarity bias, as exemplified in Figure \ref{fig:tree}, we argue that the encoder requires fine--tuning for bias mitigation. We hypothesize that $S_k$ nodes (i.e., premises that are similar to their query but irrelevant as an explanatory component), paired with their parent nodes, can act as informative negative samples that have the potential of correcting the tendency of the pre--trained encoder models to generate similar embeddings for lexically or semantically similar inputs. Such pairs can be stored alongside true positives over multiple iterations for fine--tuning the model (e.g., steps $3'$ for positives and $3''$ for negatives in Figure \ref{fig:approach}).

\begin{algorithm}[t]
	\caption{Entailment trees generation using Active Entailment Encoding}
	\begin{flushleft}
	\textbf{Input}: a hypothesis set to explain $\mathcal{H}$, premise index $\mathbb{I}$, pre--trained encoder $\mathbb{E}$, number of neighbours to retrieve K.\\
	\textbf{Output}: positive example set $\mathcal{P}$, i.e., tree branches, negative example set $\mathcal{N}$ for fine--tuning.
	\end{flushleft}
	\begin{algorithmic}[1]
		\Function{AE--Enc}{}
		    \State $\mathcal{P} \gets \{\}$; $\mathcal{N} \gets \{\}$
		    \ForAll{$H_i \in \mathcal{H}$}
		        \State $(P, N) \gets \textsc{ACS}(H_i, \mathbb{I}, \mathbb{E}, K)$
		        \State $\mathcal{P} \gets \mathcal{P} \cup P$
		        \State $\mathcal{N} \gets \mathcal{N} \cup N$
		    \EndFor
		   \State \Return $\mathcal{P, N}$
		\EndFunction
	\end{algorithmic}
\label{alg:AE-Enc}
\end{algorithm}

Algorithm \ref{alg:AE-Enc} further describes the AE--Enc procedure in practice. Input index $\mathbb{I}$ is a nearest neighbour data structure (e.g., a FAISS index) containing all the embeddings of premises in corpus $\mathcal{C}$, obtained using a pre--trained bi--encoder $\mathbb{E}$. Then, every input hypothesis is subject to retrieval based on a procedure that returns a set of positive and negative example pairs at line $4$. The positive pairs can also represent candidate tree branches while the negative pairs can be used in conjunction with the former to fine--tune encoder $\mathbb{E}$. We now describe the Active Contrastive Sampling ($\textsc{ACS}$) pair retrieval procedure (from line $4$).

\subsection{Active Contrastive Sampling}

The upper section of Figure \ref{fig:approach} depicts our proposed active strategy for generating positive and hard negative examples to reduce the similarity bias during fine--tuning. Algorithm \ref{alg:acs} defines our proposed Active Contrastive Sampling (ACS) method. Overall, the function is applied recursively for a pre--defined number of steps or until none of the retrieved candidates explain the query. Inputs $\mathbb{I}$ and $\mathbb{E}$ are the same index and encoder passed from Algorithm \ref{alg:AE-Enc}. Function $\mathsf{lookup}$ from line $4$ denotes a top--K ranked retrieval from $\mathbb{I}$. Function $\mathsf{explains}$ from line $5$ is a boolean decision function for the role of each retrieved candidate in explaining $Q$. This function can either be assigned to the user or to an automatic process that checks against a given set of known explanation trees.

\begin{algorithm}[t]
	\caption{Active Contrastive Sampling for similarity bias mitigation.}
	\begin{flushleft}
	\textbf{Input}: a query node $Q$ (i.e., hypothesis or intermediate node), premise index $\mathbb{I}$, encoder $\mathbb{E}$, number of neighbours to retrieve $K$.\\
	\textbf{Output}: positive example subset $P$, i.e., tree branches, negative example subset $N$ for fine–tuning.
	\end{flushleft}
	\begin{algorithmic}[1]
		\Function{ACS}{}
		    \State $P \gets \{\}$; $N \gets \{\}$
	        \State $emb \gets \mathbb{E}(Q)$
	        \ForAll{$P_i \in \mathbb{I}.\mathsf{lookup}(emb, K)$}
	            \If{$P_i.\mathsf{explains}(Q)$}
	                \State $P \gets P \cup \{(Q, P_i)\}$
	                \State $(P', N') \gets \textsc{ACS}(P_i, \mathbb{I}, \mathbb{E}, K)$
	                \State $N \gets N \cup N'$
		            \Else{} \State $N \gets N \cup \{(Q, P_i)\}$
		        \EndIf
		    \EndFor
		   \State \Return $P, N$
		\EndFunction
	\end{algorithmic}
\label{alg:acs}
\end{algorithm}

Algorithm \ref{alg:acs} describes a simple, yet powerful method for annotating fine--tuning contrastive samples. In contrast to typical sample selection strategies in contrastive learning (e.g., such as the ones used in \cite{Karpukhin2020DensePR}) that often rely on random, in--batch or BM25--based negative selection, ACS provides \textit{hard negatives} (i.e., highest scoring negative samples under the encoder model) specifically focused on semantic bias mitigation. At the same time, the user--labeled positives retrieved by Algorithm \ref{alg:acs} denote candidate tree branches.

\subsection{Fine--tuning bi--encoders} \label{subsec:finetune}

Having described the data generation procedure, we now describe the fine--tuning process itself. Here, we note that Algorithm \ref{alg:AE-Enc} can be applied for a pre--defined number of iterations\footnote{In practice, we have noticed that, in case of the Entailment Bank dataset, four iterations are sufficient since past that point the encoder exhibits an overfitting behavior.} to create a fine--tuning dataset. With that in hand, model $\mathbb{E}$ can be improved and the entire process restarted with only the annotated tree branches $\mathcal{P}$ being considered this time (i.e., no further hard negatives required). Following this approach and the training strategy proposed in \cite{Karpukhin2020DensePR}, two SBERT bi--encoders have been fine--tuned for query (i.e., hypothesis) and context (i.e., premise) encoding, \textit{viz.}, $\mathbb{E}_h$ and $\mathbb{E}_p$ respectively. We have experimented with different settings: the weights of both encoders are optimized during fine--tuning or only $\mathbb{E}_h$'s weights are optimized during fine--tuning. We report the results in Section \ref{sec:exp}. The latter approach has the advantage of a fixed FAISS index instead of having to recreate it after fine--tuning.

Given the two encoders and a collection of positive and negative examples of the form $(h_i, p_j^+) \in \mathcal{P}$ and $(h_i, p_j^-) \in \mathcal{N}$, respectively, a cosine score--based triplet margin loss ($TML$) \cite{Balntas_triplet16} can been used for training, as defined in Equation \ref{eq:BERT_finetuning}. Equation \ref{eq:cos} defines the cosine similarity score between embeddings, while Equation \ref{eq:tml} defines the triplet margin loss used during fine--tuning. Alternatively, supervised contrastive loss ($SCL$) \cite{Hadsell_contrastive06} can be used and we compare the two options in Section \ref{sec:exp}.

\begin{subequations}
\begin{equation}
    cos(h_i,p_j)=\frac{\mathbb{E}_h(h_i)^T~\mathbb{E}_p(p_j)}{\|\mathbb{E}_h(h_i)\|*\|\mathbb{E}_p(p_j)\|}
\end{equation} \label{eq:cos}

\begin{equation}
\begin{split}
    TML(h_i, p_j^+, p_j^-) &= \\
    \max(cos(h_i, p_j^+) - cos(h_i, p_j^i) + m, 0)
\end{split}
\end{equation} \label{eq:tml}
\label{eq:BERT_finetuning}
\end{subequations}
where $m$ is a \textit{margin} hyperparameter commonly used with triplet or contrastive losses. Intuitively, these loss functions push the members of dissimilar pairs to be further apart compared with the members of similar pairs by at least the margin value.

\subsection{Regularization} \label{subsec:reg}

The purpose of ACS is to create a balanced fine--tuning dataset that can be used to train a bi--encoder to recognize both the sentence similarity and explanatory entailment components for explanation generation purposes. In practice however, construing all lexically similar premises that are irrelevant for hypothesis explanation as negative samples can lead to negative overfitting that manifests by leaving out explanation--relevant premises that are also lexically similar. Thus, the encoders become biased towards explanatory entailment - the other end of the bias spectrum addressed in this paper. We avoid this overfitting by adding three regularization terms to the loss function\footnote{A similar regularization scheme with only two factors, i.e., for the hypothesis and for the example premise, could be devised for SCL.}, as shown in Equation \ref{eq:BERT_finetuning_regularisation}. 

\begin{equation}
\begin{split}
	TML(h_i, p_j^+, p_j^-) &= \\
    \max(cos(h_i, p_j^+) - cos(h_i, p_j^-) + m, 0) \\
	+\alpha\|\mathbb{E}_s(h_i)-\mathbb{E}_h(h_i)\|^2_2\\
	+\alpha\|\mathbb{E}_s(p_j^+)-\mathbb{E}_p(p_j^+)\|^2_2\\
	+\alpha\|\mathbb{E}_s(p_j^-)-\mathbb{E}_p(p_j^-)\|^2_2
	\end{split}
	\label{eq:BERT_finetuning_regularisation}
\end{equation}
where $\alpha$ is a regularization weight hyperparameter.

Intuitively, we propose a regularization scheme based on a \textit{fixed sentence/passage encoder}, i.e., a pre--trained Sentence--BERT $\mathbb{E}_s$ in Equation~\ref{eq:BERT_finetuning_regularisation}. This fixed encoder is not fine--tuned and, therefore, biased towards lexical and semantic similarity. Thus, we counteract entailment bias by re--introducing sentence similarity bias in a controlled manner.

\section{Experiments} \label{sec:exp}

\begin{table*}[t]
	\centering
	\caption{MAP, NDCG, and Hit@K results on EB test split. SCL means Supervised Contrastive Loss. TML means Triplet Margin Loss, best performing results are tagged by bold font and best baseline results are tagged by underline.}
	\label{tab:evaluation_results}
	\scalebox{0.78}{
		\begin{tabular}{@{}lcc|ccccc|ccccc@{}}
			\toprule
% 			& \multicolumn{2}{c|}{25 facts} & \multicolumn{8}{c}{All facts}\\
% 			\midrule
			& & & \multicolumn{5}{c}{NDCG@k} & \multicolumn{5}{c}{Hit@k} \\
			& MAP & NDCG & 10 & 20 & 30 & 40 & 50 & 10 & 20 & 30 & 40 & 50\\
			\midrule
			TF-IDF & 0.335 & 0.525 & 0.446 & 0.476 & 0.487 & 0.493 & 0.496 & 0.628 & 0.726 & 0.766 & 0.793 & 0.810\\
			BERT-base-uncased & 0.067 & 0.206 & 0.098 & 0.105 & 0.108 & 0.111 & 0.113 & 0.127 & 0.148 & 0.161 & 0.174 & 0.181\\
			DPR & 0.243 & 0.438 & 0.331 & 0.363 & 0.376 & 0.384 & 0.388 & 0.464 & 0.566 & 0.612 & 0.648 & 0.663\\
			SimCSE & 0.324 & 0.517 & 0.436 & 0.463 & 0.476 & 0.481 & 0.485 & 0.615 & 0.702 & 0.749 & 0.772 & 0.788\\
			T5-base+enc-mean & 0.268 & 0.459 & 0.368 & 0.390 & 0.402 & 0.408 & 0.413 & 0.505 & 0.577 & 0.624 & 0.649 & 0.669\\
			SBERT & 0.356 & 0.546 & 0.470 & 0.500 & 0.511 & 0.517 & 0.521 & 0.666 & 0.763 & 0.804 & 0.828 & 0.845\\
			\hline
			T5-base+enc-mean+SCL & 0.348 & 0.534 & 0.452 & 0.476 & 0.488 & 0.495 & 0.499 & 0.599 & 0.678 & 0.723 & 0.752 & 0.770\\
			SBERT+SCL Single-enc & \underline{0.388} & \underline{0.574} & \underline{0.505} & \underline{0.532} & \underline{0.543} & \underline{0.549} & \underline{0.552} & 0.701 & 0.787 & 0.826 & 0.852 & 0.869\\
			~~~(VS SBERT $\uparrow$) & 9.0\% & 5.1\% & 7.4\% & 6.4\% & 6.2\% & 6.1\% & 5.9\% & - & - & - & - & -\\
			SBERT+TML Single-enc & 0.376 & 0.563 & 0.491 & 0.525 & 0.536 & 0.543 & 0.545 & 0.708 & \underline{0.818} & \underline{0.862} & \underline{0.888} & \underline{0.901}\\
			~~~(VS SBERT $\uparrow$) & - & - & - & - & - & - & - & - & 7.2\% & 7.2\% & 7.2\% & 6.6\%\\
			SBERT+TML Siamese-enc & 0.372 & 0.560 & 0.491 & 0.522 & 0.532 & 0.538 & 0.542 & \underline{0.713} & 0.814 & 0.851 & 0.878 & 0.896\\
			~~~(VS SBERT $\uparrow$) & - & - & - & - & - & - & - & 7.1\% & - & - & - & -\\
			SBERT+TML Dual-enc & 0.366 & 0.555 & 0.483 & 0.515 & 0.527 & 0.534 & 0.537 & 0.710 & 0.811 & 0.858 & 0.886 & 0.900\\
			\hline
			AE--Enc & 0.447 & 0.617 & 0.550 & 0.581 & 0.591 & 0.597 & 0.600 & 0.724 & 0.824 & 0.862 & 0.887 & 0.902\\
			AE--Enc no regularisation & 0.346 & 0.527 & 0.439 & 0.468 & 0.481 & 0.488 & 0.493 & 0.606 & 0.698 & 0.748 & 0.775 & 0.798\\
			Iterative AE--Enc & \textbf{0.451} & \textbf{0.622} & \textbf{0.562} & \textbf{0.589} & \textbf{0.599} & \textbf{0.605} & \textbf{0.608} & \textbf{0.758} & \textbf{0.842} & \textbf{0.881} & \textbf{0.904} & \textbf{0.920}\\
			~~~(VS best baseline $\uparrow$) & 16.2\% & 15.3\% & 11.3\% & 10.7\% & 10.3\% & 10.2\% & 10.1\% & 6.3\% & 2.9\% & 2.2\% & 1.8\% & 2.1\% \\
			\bottomrule
		\end{tabular}
		}
\end{table*}

Our proposal in this paper, AE--Enc, targets human--in--the--loop explanation tree construction scenarios. Our central hypothesis is that AE--Enc in general and ACS in particular can ensure a balanced fine--tuning dataset for generating trees and, consequently, can lead to embedding spaces that capture both types of explanatory relationships better than the alternatives. To validate this hypothesis, in this section, we first perform an extrinsic evaluation of AE--Enc by comparison to several baselines, aiming to show the beneficial impact of unbiased encoder models on the construction of entailment trees. We then analyze our methods further by performing an intrinsic evaluation where our desideratum is to show that the premise embedding space generated by a fine--tuned bi--encoder is equally defined by lexical or semantic similarity and entailment relationships.

\subsection{Experimental setup}

We use the Entailment Bank (EB) \cite{Dalvi2021ExplainingAW} as the \textbf{dataset} for both fine--tuning and evaluation. From EB's trees, we extract all the parent--child node pairs to generate $8,801$ positive hypothesis--premise examples. EB also provides negative examples (i.e., distractors) for each explanation tree. We use these distractors in conjunction with the positive pairs to generate a triplet store that acts as the initial gold standard and we use it to simulate human selection. In other words, function $\mathsf{explains}$ from Algorithm \ref{alg:acs} evaluates the explainability relevance of a given candidate with respect to the gold standard. Overall, we generate $167,082$ gold (hypothesis, premise, distractor) triplets.

With respect to hyperparameter configuration we set both the margin value $m$ from Equation \ref{eq:tml} and the regularization factor $\alpha$ to $0.1$. We train our encoder models with different combinations of batch sizes from $\{32, 64, 128\}$ and learning rates from $\{1e-5, 3e-5, 5e-5\}$ and report the results of the best--performing combination on the \textit{dev} data set. The best--performing setting was batch size $=32$, learning rate $=1e-5$, and models fine--tuned for $5$ epochs.

\subsection{Reported measures}

Since we frame premise--selection as a ranked retrieval task, in our extrinsic analysis we employ two widely used metrics in ranking evaluation and an additional metric specific to simulated human selection.

\begin{itemize}[leftmargin=*, nosep]
    \item \textit{Mean average precision} (MAP) is used to measure overall prediction quality of the ranked results. 
    \item \textit{Normalized Discounted Cumulative Gain at K} (nDCG@K) is used to measure the ranking quality in the top--K retrieved results - with various values of K reported. 
    \item \textit{Hit at K} (Hit@K) is used to evaluate the simulation of human selection by measuring the ratio of explanatory--relevant premises in the top--K results for a given query - with various values of K reported. 
\end{itemize}

\subsection{Baselines}

For the extrinsic evaluation, we use several baseline models to show the impact of similarity bias on the retrieval--based premise--selection task:

\begin{itemize}[leftmargin=*, nosep]
    \item \textit{TF--IDF} creates vector representations from tf--idf scores.
    \item DPR (Dense Passage Retrieval) \cite{Karpukhin2020DensePR} uses two separate encoders, for query and premises, to perform passage retrieval at scale.
    \item \textit{Simple Contrastive Learning of Sentence Embeddings} (SimCSE) \cite{Gao2021SimCSESC} is a simple BERT--based bi--encoder trained on both dropout--based augmented samples and NLI samples.
    \item \textit{Sentence--T5} \cite{Ni2022SentenceT5SS} is a T5 \cite{Raffel-T519} --based sentence encoder originally proposed for generative tasks. We use it with mean pooling.
    \item \textit{Sentence--BERT} (SBERT) in various settings: as a simple Siamese encoder (i.e., query and premise encoders share weights), as a Dual encoder (i.e., query and premise encoders are both updated during fine--tuning), as a Single encoder (only the query encoder is updated during fine--tuning)
\end{itemize} .

We use the above baseline models in various settings defined by the loss function used: SCL \textit{vs.} TML, or by the architecture type used: Siamese \textit{vs.} Single \textit{vs.} Dual encoders. We use the following models pre--trained without any fine--tuning: BERT-base-uncased, DPR, SimCSE, SBERT, T5. In the results, we denote their fine--tuned versions by specifying the loss type: SCL or TML. For each of the baselines, we generate embeddings for each hypothesis/premise in the EB corpus and index them using FAISS for fast retrieval.

We evaluate the baselines against three fine--tuned configurations of our proposed method: AE--Enc with and without regularization and a one time active augmentation of the fine--tuning data (i.e., Algorithm \ref{alg:acs} is applied once and the results are added to the fine--tuning data) or AE--Enc with regularization and iterative augmentation of the fine--tuning data  (i.e., Algorithm \ref{alg:acs} is applied multiple times). In the latter case, we have observed that after four iterations the generated fine--tuning dataset leads to encoder overfitting.

\subsection{Extrinsic evaluation results} \label{subsec:extrinsic}

Table \ref{tab:evaluation_results} reports the values of MAP, NDCG(@K) and Hit@K obtained on EB's test data. The first two columns correspond to entire candidate corpus of 16,471 unique candidate premises.

The first 6 rows in the table correspond to pre--trained models without any domain--specific fine--tuning. It can be noted that, overall, the retrieval grounded on TF--IDF and SBERT perform best in this case. Comparing SBERT+SCL Single--enc and SBERT+TML Single--enc shows a marginal difference. Similarly, when comparing SBERT+TML Single--enc, SBERT+TML Siamese--enc, and SBERT+TML Dual--enc, the differences are marginal as well. 

Overall, among the fine--tuned models initialized with SBERT, SBERT+SCL Single--enc performs best: 9.0\% MAP improvement and 5.1\% NDCG improvement over pre--trained SBERT. We also observe a premise ranking improvement (i.e., in NDCG@K) over pre--trained SBERT of 5.9\% to 7.4\%. On the evaluation at @K SBERT+TML Siamese-enc and SBERT+TML Single-enc are the best performing baselines, with a 6.6\% to 7.2\% improvement over pre--trained SBERT. Overall, although the MAP and NDCG improvements characterizing the fine--tuned baselines are noticeable, it is questionable if they justify the effort and costs often necessary to obtain the fine--tuning dataset in domain--specific scenarios. We observe that these reduced improvements brought on by the fine--tuned models suggests that the fine--tuning process does not enhance the ability of the models to identify other types of hypothesis--premise relationships beyond lexical or semantic similarities, both of which are already captured by the pre--trained version. Conversely, our proposed method, AE--Enc, contributes with additional entailment--specific signal. Specifically, Iterative AE--Enc with regularization (i.e., Iterative AE--Enc) leads to an improvement in MAP and NDCG of 16.2\% and 15.3\%, respectively, compared with SBERT+SCL Single-enc. These results support our claim that entailment--specific relationship awareness can improve the premise-selection accuracy. 

Performance improvements are also observed in the case of Hit@10 where Iterative AE--Enc leads to the retrieval of up to 6.2\% more relevant premises than the best performing alternatives SBERT+TML Siamese--enc/Single--enc.

When comparing different AE--Enc configurations (i.e., the last three rows of Table \ref{tab:evaluation_results}) we observe that applying Algorithm \ref{alg:acs} iteratively and employing regularization leads to the best overall results on all reported measures. The results suggest that, as hypothesized in Section \ref{subsec:reg}, AE--Enc without regularization suffers from entailment bias and becomes ineffective at identifying lexical similarity. In fact, the results corresponding to AE--Enc without regularization are even worse than those corresponding to pre--trained SBERT, further supporting our research hypothesis that both similarity and entailment relationships are necessary in generating explanation trees.

\subsection{Intrinsic evaluation}

\begin{figure*}[t]
    \centering
     \begin{subfigure}{0.49\textwidth}
        \includegraphics[width=\textwidth]{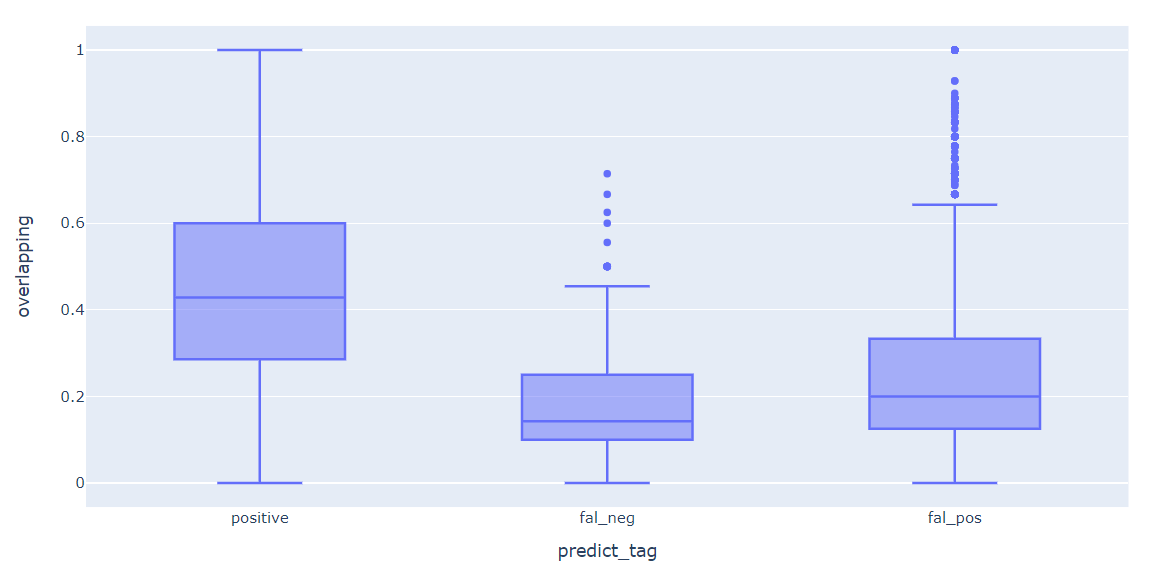}
        \caption{Token overlap degrees of TPs, FNs and FPs groups.}
        \label{fig:overlapping_positive_negative}
     \end{subfigure}
     \begin{subfigure}{0.49\textwidth}
        \includegraphics[width=\textwidth]{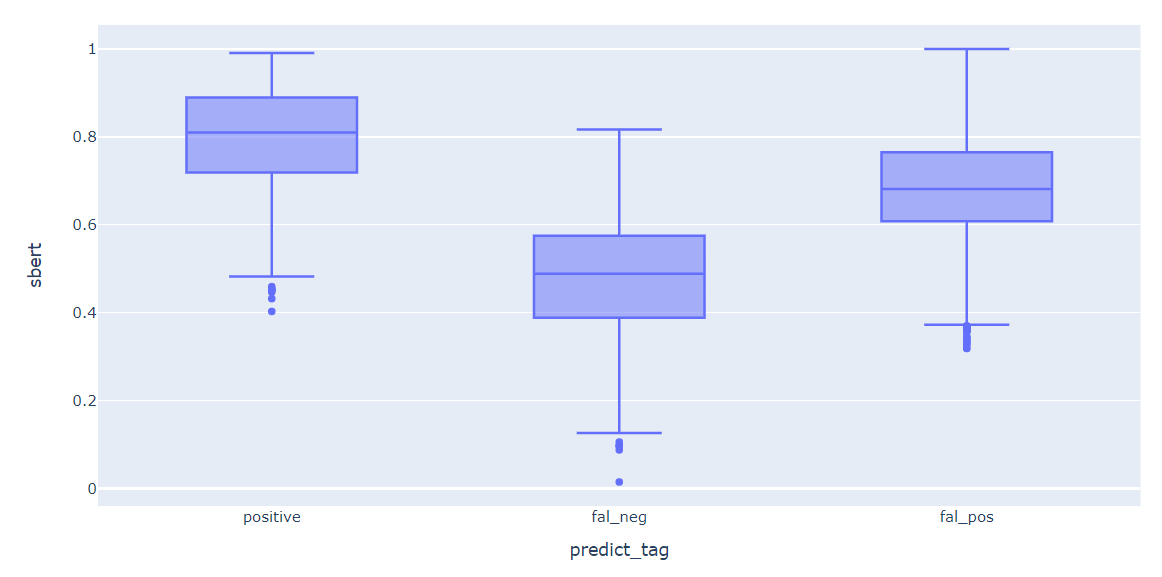}
        \caption{Cosine similarity of TPs, FNs and FPs groups - embeddings produced by pre--trained SBERT.}
        \label{fig:cosineSim_positive_negative}
     \end{subfigure}
     \begin{subfigure}{0.49\textwidth}
        \includegraphics[width=\textwidth]{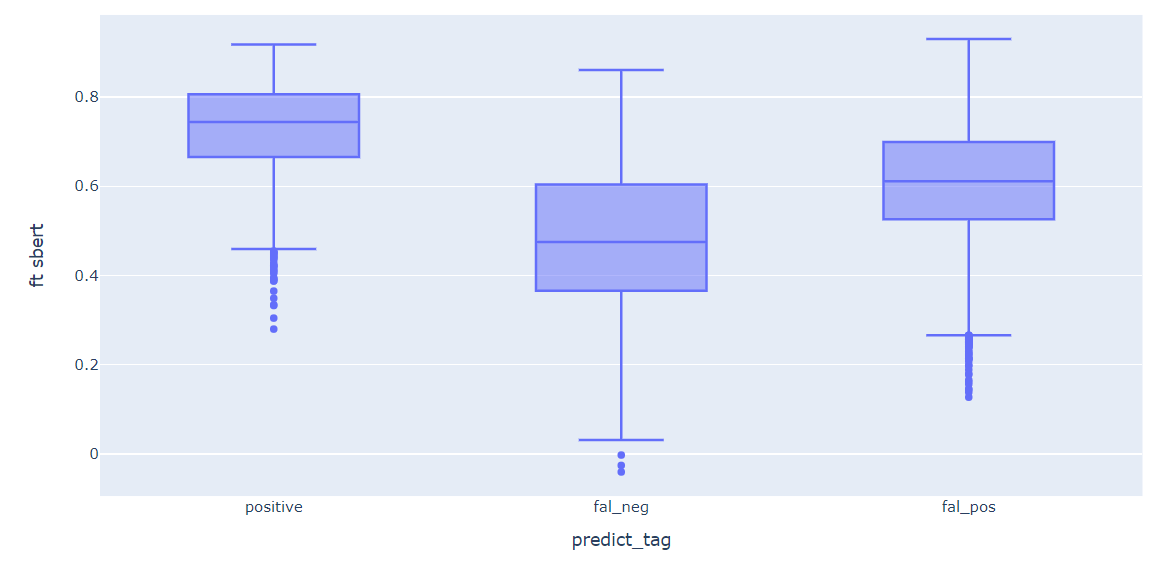}
        \caption{Cosine similarity of TPs, FNs and FPs groups - embeddings produced by SBERT fine--tuned on EB data.}
        \label{fig:ft-sbert_positive_negative}
     \end{subfigure}
     \begin{subfigure}{0.49\textwidth}
        \includegraphics[width=\textwidth]{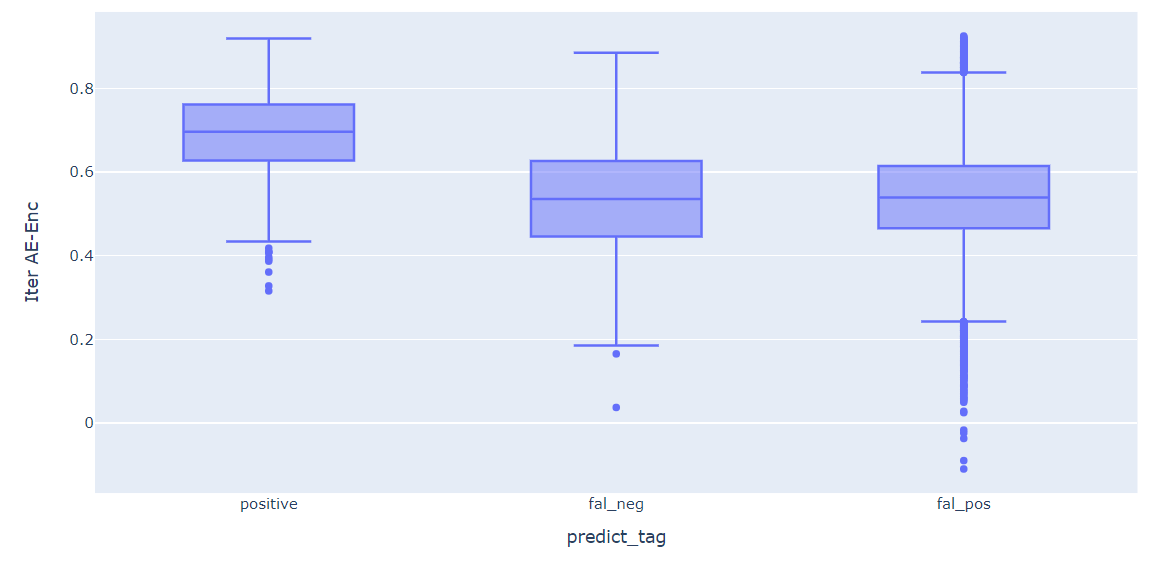}
        \caption{Cosine similarity of TPs, FNs and FPs groups - embeddings produced by SBERT fine--tuned on data generated by Iterative AE--Enc.}
        \label{fig:iter-ae-enc_positive_negative}
     \end{subfigure}
     \caption{Similarity analysis showing the bias associated with sentence bi--encoders. Num. TPs = 1745; Num. FNs = 541, Num. FPs = 19935.}
    \label{fig:overlapping_failure}
\end{figure*}

To further analyze the potential of Iterative AE--Enc to overcome the sentence similarity bias, we investigate the embedding similarity distributions for true positives (TPs), false positives (FPs) and false negatives (FNs). For comparison purposes, TP, FP and FN labels are fixed and assigned by a pre--trained SBERT. We only use the fine--tuned versions to recreate the representations of each sentence and analyze how the similarity between the resulting embeddings changes in each group. The aim in this section is, therefore, to analyze how does the fine--tuning process modifies the resulting premise embedding space, relative to the hypothesis--premise pairs deemed TPs, FPs or FNs by the biased, pre--trained SBERT.

We start by using a pre--trained SBERT encoder to encode the entire premise corpus $\mathcal{C}$ of the Entailment Bank\footnote{16471 unique candidate premises.} and index it using FAISS. We use the resulting index to perform top--K rank retrieval\footnote{We used K=20 in practice.} using all (parent node, child node) pairs from the test entailment trees of EB as evaluation: i.e., if the child node is in the top--K retrieved result for the parent node then it is marked as a TP and as a FN otherwise. Any other retrieved node that is not part of the same tree in the test data is marked as a FP.

The results of our similarity analysis are reported in Figure \ref{fig:overlapping_failure}. Figure \ref{fig:overlapping_positive_negative} reports the overlap as the Jaccard similarity computed at token level between the sentences of TPs (i.e., positives), FNs and FPs. The figure reiterates the behavior exemplified in Figure \ref{fig:tree}: non--relevant premises for explanation purposes are deemed similar by the pre--trained bi--encoder because of their high lexical overlap with the query. The same behavior is shown in Figure \ref{fig:cosineSim_positive_negative} where the cosine similarity between the query and candidate embeddings generated by the pre--trained bi--encoder is reported. Once again, FPs are characterized by higher similarity than FNs. 

When we analyze the similarity space defined by a version of the model fine--tuned on Entailment Bank data, in Figure \ref{fig:ft-sbert_positive_negative}, we can observe an overlap between the similarity distributions of FNs and FPs - a sign that the fine--tuning process has the potential of overcoming the bias towards similar sentences if fueled by informative examples. In this paper, we argue that the similarity bias can be further reduced by supplying the bi--encoders with better positive and negative example pairs that characterized broader entailment relationships between hypothesis and premise. We prove this hypothesis in Figure \ref{fig:iter-ae-enc_positive_negative} that describes the case where the embeddings result from a Sentence--BERT model fine--tuned on data generated by Iterative AE--Enc. By comparison with Figures \ref{fig:cosineSim_positive_negative} and \ref{fig:ft-sbert_positive_negative} it can be seen that the newly fine--tuned model balances the degree of similarity of FNs and FPs. In other words, the proximity of hypothesis and candidate premises in the embedding space is less influenced by the lexical overlap. This is consistent with and explains the AE--Enc's superior performance reported in Section \ref{subsec:extrinsic}. 

\section{Conclusion}

In this paper, we propose an active learning strategy called Active Entailment Encoding (AE--Enc) that uses a sampling strategy specialized for explanatory entailment, ACS, to generate fine--tuning data for Transformer--based entailment tree construction. We aim to balance the bias towards lexical and semantic similarity characteristic to state--of--the--art sentence/passage encoders. Our strategy is proposed in the context of human--in--the--loop explanation tree construction, a task characterized by the need for balancing both similarity and entailment relationships. In achieving this balance, using ACS, we augment fine--tuning data for retrieval models with the necessary positive and negative examples to increase entailment--awareness. Our method is primarily focused on the identification of hard negatives that can counteract the sentence similarity bias while preserving entailment--specific signal. We further control the influence of these hard negatives by proposing a simple regularization scheme. 

Our extrinsic evaluation shows that, indeed, regularized AE--Enc achieves best premise selection results for entailment trees generation when compared against several baselines. This is further supported by an intrinsic analysis that shows the ability of our proposal to moderate the influence the lexical similarity between hypothesis and explanatory premises has on the proximity of the two in the embedding space.

\bibliography{anthology,custom}
\bibliographystyle{acl_natbib}

% \appendix

% \section{Example Appendix}
% \label{sec:appendix}

% This is an appendix.

\end{document}